\documentclass[11pt]{article}

\usepackage[preprint]{acl}

\usepackage{times}
\usepackage{latexsym}

\usepackage[T1]{fontenc}

\usepackage{booktabs}

\usepackage[utf8]{inputenc}

\usepackage{microtype}

\usepackage{inconsolata}

\usepackage{graphicx}
\usepackage{amsmath}
\usepackage{amssymb}

\title{How Much Information Can a Vision Token Hold? A Scaling Law for Recognition Limits in VLMs}

\author{
    \textbf{Shuxin Zhuang}\textsuperscript{1,4}\thanks{ \ Equal contribution.},
    \textbf{Zi Liang}\textsuperscript{2}\footnotemark[1],
    \textbf{Runsheng Yu}\textsuperscript{3},
    \textbf{Hongzong Li}\textsuperscript{3}\thanks{ \ Corresponding author.}, \\
    \textbf{Rong Feng}\textsuperscript{1,4},
    \textbf{Shiqin Tang}\textsuperscript{4},
    \textbf{Youzhi Zhang}\textsuperscript{4} \\
    \textsuperscript{1}City University of Hong Kong \quad
    \textsuperscript{2}The Hong Kong Polytechnic University \\
    \textsuperscript{3}The Hong Kong University of Science and Technology \\
    \textsuperscript{4}Centre for Artificial Intelligence and Robotics, Chinese Academy of Sciences \\
    \small \texttt{\{shuxin.zhuang, rongfeng3-c\}@my.cityu.edu.hk}, \texttt{zi1415926.liang@connect.polyu.hk} \\
    \small \texttt{runshengyu@gmail.com}, \texttt{lihongzong@ust.hk}, \texttt{\{shiqin.tang, youzhi.zhang\}@cair.cas.org.hk}
}

\begin{document}
\maketitle
\begin{abstract}
Recent vision-centric approaches have made significant strides in long-context modeling. Represented by DeepSeek-OCR, these models encode rendered text into continuous vision tokens, achieving high compression rates without sacrificing recognition precision. However, viewing the vision encoder as a lossy channel with finite representational capacity raises a fundamental question: what is the information upper bound of visual tokens? To investigate this limit, we conduct controlled stress tests by progressively increasing the information quantity (character count) within an image. We observe a distinct phase-transition phenomenon characterized by three regimes: a near-perfect \textit{Stable Phase}, an \textit{Instability Phase} marked by increased error variance, and a total \textit{Collapse Phase}. We analyze the mechanical origins of these transitions and identify key factors. Furthermore, we formulate a probabilistic scaling law that unifies average vision token load and visual density into a latent difficulty metric. Extensive experiments across various Vision-Language Models demonstrate the universality of this scaling law, providing critical empirical guidance for optimizing the efficiency-accuracy trade-off in visual context compression.
\end{abstract}

\section{Introduction}

The escalating demand for long-context understanding in Large Language Models (LLMs) is constrained by the quadratic computational complexity of self-attention mechanisms. To mitigate this bottleneck, context compression has emerged as a critical research frontier. While approaches such as efficient attention \cite{yang2024gated, peng2025rwkv, chen2025minimax}, retrieval-augmented generation (RAG) \cite{laban2024summary, yu2025memagent}, and text-level compression \cite{liu2025context} have made strides, relying solely on the text-only modality inevitably discards critical spatial and structural semantics (e.g., tabular alignments, hierarchical indentation, and floating layouts). Inspired by DeepSeek-OCR \cite{wei2025deepseekocr}, which encode rendered text images into continuous vision tokens to reconstruct textual content, the vision encoder can also serve as a mechanism for information compression. By encoding information into vision tokens, this approach not only maximally preserves spatial-structural semantics but also alleviates the computational burden imposed by long input sequences, suggesting a paradigm shift towards vision-centric language modeling, potentially replacing discrete text tokenization with continuous visual representations for all data modalities.

Existing studies have validated the practical potential of this paradigm, reporting high decoding precision at moderate compression ratios. However, these evaluations report end-to-end precision at selected compression ratios on document benchmarks, making it difficult to distinguish whether failures stem from the information capacity limits of the vision tokens or simply from rendering artifacts. This introduces a critical challenge: \textbf{What is the upper bound of semantic information that can be compressed into vision tokens?} In current Vision-Language Model (VLM) architectures, images are projected into a sequence of vision tokens serving as inputs to the LLM. Since the computational complexity of self-attention is directly determined by the number of vision tokens, minimizing this count is crucial for computational efficiency. Unlike discrete text tokenization (e.g., BPE) which guarantees lossless reconstruction, the vision encoder functions as a lossy compression channel with a finite capacity. This creates a critical trade-off between computational efficiency and representational capacity: while minimizing the vision token count reduces the self-attention overhead, it forces each token to encode a denser semantic payload, potentially exceeding the information capacity of a vision token. Consequently, this raises an unexplored question: What is the quantitative relationship between the information quantity within an image and the number of vision tokens used to encode it? To ensure reliable reconstruction, where does the capacity limit of visual tokens lie?

To address these questions, we investigate the effective capacity of visual tokens under a fixed token budget. We focus on the task of decoding dense pure text rendered as images, as this setting maximizes the information quantity within an image. We synthesize text images across varying semantic and typographic conditions to serve as a controlled testbed. By constraining the vision token budget and controlling information quantity, we analyze the reconstruction behavior as the rendered content extends. Our experiments reveal a phase-transition phenomenon (illustrated in Figure~\ref{fig:scatter_plot_main}) where, instead of degrading linearly with increasing information quantity, performance follows a three regime behavior: a near-perfect \textbf{Stable Phase}, an \textbf{Instability Phase} where errors fluctuate dramatically under comparable loads, and a \textbf{Collapse Phase} once beyond a critical boundary that we term the \textbf{Hard Wall}. We further distinguish two underlying factors driving these transitions: (1) instability arising from \emph{spatial alignment sensitivity} in Vision Transformer (ViT) patch partitioning; and (2) an irreversible collapse driven by an \emph{information capacity limit}, occurring when the information quantity exceeds the representational power of the vision tokens.

Beyond characterizing these phase transitions, we develop a probabilistic scaling-law model that captures the effects of \emph{average vision token load} and \emph{visual density} on recognition performance, thereby providing a practical tool for estimating maximum information quantity under a fixed vision token budget. Finally, we validate the generalizability of this probabilistic scaling law across representative modern VLM architectures, suggesting that the observed phase transition behavior reflects a broader property of current vision tokenization and ViT-based encoding.

Our contributions are summarized as follows:
\begin{itemize}
    \item \textbf{Discovery of a Scale-Invariant Phase Transition:} We identify a phase-transition phenomenon in visual token reconstruction characterized by distinct Stable, Instability, and Collapse phases. Crucially, we observe that the transition width is \textit{resolution-independent}: the instability phase consistently spans an approximate $\mathbf{2.2\times}$ range in text length.
    \item \textbf{Two distinct failure mechanisms:} We distinguish between instability caused by spatial alignment sensitivity in ViT patch partitioning and irreversible collapse driven by the information capacity limit of vision tokens.
    \item \textbf{Probabilistic scaling law:} We formulate a model that correlates average vision token load and visual density with recognition performance to estimate the maximum information capacity under a fixed budget.
\end{itemize}

\section{Related Work}

\subsection{Long-Context Compression}
Scaling the context window of LLMs to handle million-level tokens presents significant challenges in terms of memory and computational overhead. Consequently, context compression has emerged as a critical research direction to alleviate these bottlenecks \cite{hu2025longrecipe}.

Recent studies have innovated by transcending traditional text-based token limits, exploring visual and optical compression mechanisms. \citet{wei2025deepseekocr} introduced DeepSeek-OCR. By utilizing a vision encoder to map long textual contexts into 2D optical representations, this work demonstrated the feasibility of visual mapping for enhancing memory efficiency. Following this visual-centric paradigm, \citet{cheng2025glyph} proposed Glyph, a framework that extends context windows by rendering text into images processed by VLMs. 

In contrast to cross-modal approaches, \citet{liu2025context} explored the practical limits of compression through a pure-text approach termed Context Cascade Compression (C3). This method employs a cascading architecture where a smaller LLM compresses long contexts into compact latent tokens, which are then decoded by a larger LLM. It demonstrated that this text-to-text latent compression significantly outperforms current optical methods, achieving 98\% decoding accuracy at a 20x compression ratio. Despite achieving higher compression ratios, pure text compression inevitably discards critical spatial and structural semantics.

\subsection{End-to-End Visual Document Understanding}

With the rise of Transformers, document understanding technology has increasingly evolved toward end-to-end architectures, encompassing encoder-decoder OCR \citep{li2023trocr} and document intelligence models that fuse text, layout, and visual cues \citep{xu2020layoutlm,tang2023udop}. In parallel, OCR-free paradigms have emerged in the field of document understanding, demonstrating their potential by directly generating structured outputs from document images \cite{kim2022ocr} and through broader image-to-text pretraining schemes for visually-situated language \cite{lee2023pix2struct}.

Despite rapid progress, these works primarily focus on optimizing model design and benchmark accuracy \cite{zhao2025vtcbench}, and their evaluation protocols typically emphasize task-level correctness. However, to the best of our knowledge, no work has explored exactly how much textual information each vision token can faithfully carry. This question becomes critical when vision tokens are explicitly used as a interface for LLMs \citep{wei2025deepseekocr}, which is exactly what drives us to conduct vision token capacity limit analysis.

\begin{figure}[t]
\centering
\includegraphics[width=\columnwidth]{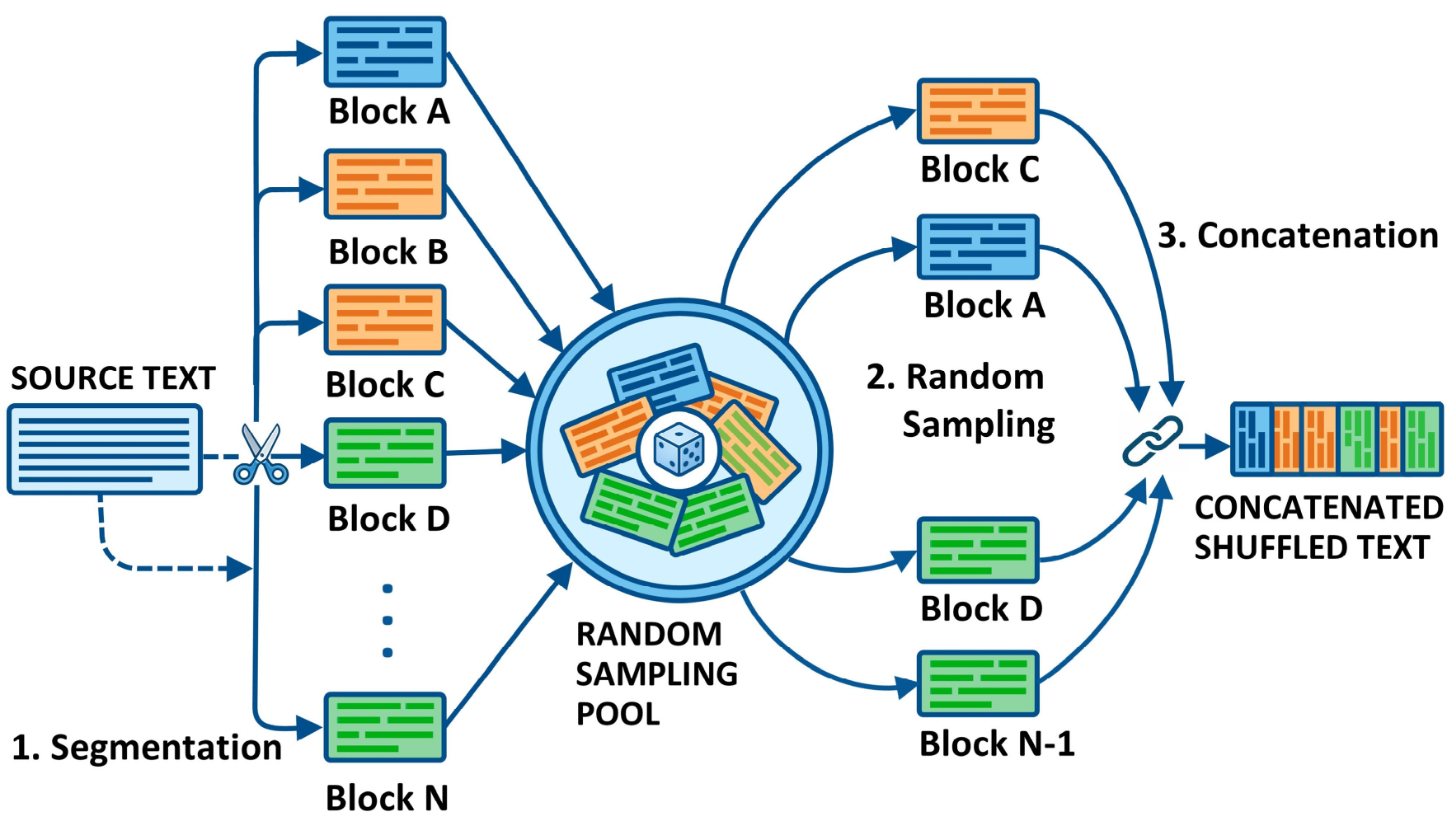}
\caption{Illustration of the Block-wise Shuffling strategy. Text blocks are segmented, randomly sampled, and concatenated to construct randomized semantic text.}
\label{fig:image_generation}
\end{figure}

\begin{figure*}[t]
\centering
\includegraphics[width=\textwidth]{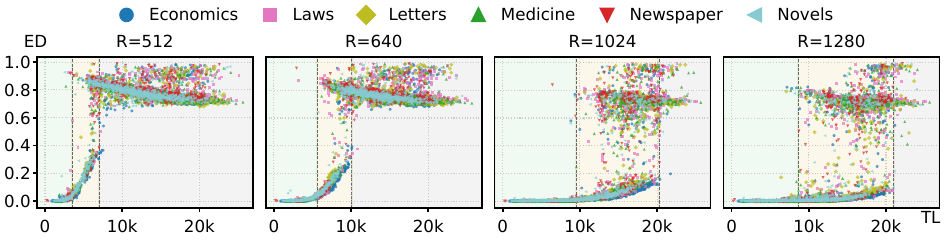}
\caption{Scatter plots of ED versus Text Length (TL) across four resolutions ($R$). Three distinct regimes (shaded) emerge across semantic domains: Stable Phase (green), Instability Phase (yellow), and Collapse Phase (gray). Higher resolutions exhibit a wider Instability Phase and shift the hard wall to the right.}
\label{fig:scatter_plot_main}
\end{figure*}

\section{Experimental Setup}
\label{sec:experimental_setup}

\subsection{Data Synthesis Pipeline}
To evaluate the vision token capacity across different semantic domains, we collected a diverse set of public domain texts from \href{https://www.gutenberg.org}{Project Gutenberg}. We selected six text categories: \textit{Novels, Laws, Economics, Medicine, Newspapers,} and \textit{Letters}. This diversity ensures that our findings are invariant to domain-specific vocabulary and syntactic structures.

Central to our design is the decoupling of visual recognition from semantic prediction. LLMs rely on strong contextual priors to predict the next token, which can mask failures in the visual perception module. To mitigate this bias, we implemented a \textit{Block-wise Shuffling} strategy. The process is illustrated in Figure~\ref{fig:image_generation}. Specifically, source texts are segmented into several discrete blocks, which are then randomly sampled and concatenated during the image generation process. This randomization of semantic order forces the model to rely solely on visual information rather than exploiting linguistic priors for next-token prediction.

We adjust typographical parameters, including font size, line spacing, and character spacing, to simulate different typographic densities within the images. Since the span of text length varies significantly, fixing the generated image width would result in extreme aspect ratios. Therefore, we employ a dynamic layout adjustment algorithm. This algorithm adjusts the text wrapping width to ensure the final image maintains an aspect ratio between 0.9 and 1.1. Detailed descriptions of text categories, corpus construction, and image rendering specifics are provided in Appendix \ref{app:data_synthesis}.

\subsection{Model Selection and Architecture}
\label{sec:model_architecture_selection}

Our study aims to investigate the relationship between the information quantity carried by an image and the capacity limit of a fixed vision token budget. To conduct this analysis, we must select a suitable VLM that satisfies the following criteria: (i) it is capable of processing images with high information quantity; (ii) the model architecture allows for control over the number of vision tokens; and (iii) the model represents the state-of-the-art performance in image content recognition. Based on these requirements, we select DeepSeek-OCR as our primary experimental model. Its core DeepEncoder employs a serial design that cascades SAM-based local window attention and CLIP-based global attention with downsampling. This mechanism compresses visual inputs into compact latent tokens and enables precise control over the token budget by adjusting model's input resolution.

To ensure that our findings describe intrinsic properties of Vision Transformers rather than artifacts specific to the hybrid architecture, we extend our evaluation to representative VLMs from alternative architectural categories defined in the \citet{wei2025deepseekocr}. Specifically, we select \textbf{InternVL3.5-8B} to represent the \textit{Tile-based High-Resolution} strategy, which typically crops images into local tiles combined with a global thumbnail to preserve details. Additionally, we include \textbf{Qwen2.5-VL-8B} as the representative of the \textit{Native Dynamic (NaViT)} strategy, utilizing varied sequence lengths without padding to naturally handle arbitrary aspect ratios.

All experiments were conducted on a server equipped with dual Intel® Xeon® Gold 6430 CPUs (2$\times$32 cores, 128 threads total) and an NVIDIA Tesla A100 GPU with 80 GB memory. Our code and experimental configurations are publicly available at \url{https://github.com/sxzhuang/Scaling-Law-for-Vision-Token}.

\subsection{Evaluation Metrics}

\paragraph{Rationale for Text-Only Evaluation}
While real-world documents often contain tables and figures, our study focuses exclusively on text-only images. This choice is driven by two considerations. First, compared to charts or geometric figures which contain whitespace and structural redundancy, pure text represents the highest information quantity per image. This enables us to effectively probe the upper bound of the vision token capacity. Second, quantitative analysis requires a controllable independent variable. Pure text allows for manipulation of information quantity via exact character counts. 

\paragraph{Edit Distance}
Given our exclusive focus on textual content, we utilize edit distance (ED)~\cite{lcvenshtcin1966binary} as the metric to quantify reconstruction quality. This metric calculates the minimum number of single-character operations (insertions, deletions, or substitutions) required to transform the model's predicted sequence into the ground truth sequence. A lower ED indicates higher recognition accuracy.

\begin{figure*}[t]
\centering
\includegraphics[width=\textwidth]{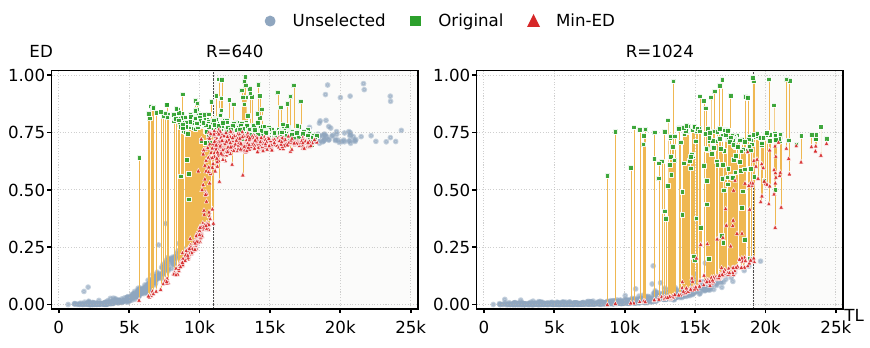}
\caption{Results of the Pixel-Shift Perturbation experiment. We plot ED against Text Length (TL) for resolutions $R=640$ and $R=1024$. The black vertical dashed line marks the Hard Wall separating Zone I from Zone II. The yellow vertical lines connect the original performance (green squares) to the minimum ED (red triangles) achieved after perturbation for the same sample. The visualization demonstrates that errors in Zone I are reversible through spatial alignment, whereas errors in Zone II persist regardless of pixel shifting.}
\label{fig:pixel_shift}
\end{figure*}

\section{From Stability to Collapse}
\label{sec:hard_wall_mechanics}

In this section, we investigate how recognition accuracy changes as the information quantity in the input image increases. We evaluate DeepSeek-OCR under four input resolutions ($R \in \{512, 640, 1024, 1280\}$) and observe that the model exhibits critical phase transitions, culminating in a sudden failure point we term the ``Hard Wall.'' The following analysis characterizes these distinct regimes and dissects the specific failure mechanisms—ranging from spatial misalignment to capacity exhaustion—that drive the transition from stability to collapse.

\subsection{Distinct Performance Regimes}
\label{sec:performance_regimes}

As shown in Figure~\ref{fig:scatter_plot_main}, we examined whether the semantic complexity of the text influences recognition accuracy. By comparing results across six distinct text categories, we observed that the reconstruction error was largely independent of the semantic domain. This confirms that under our block-wise shuffling setup, the model relies primarily on visual perception rather than linguistic priors. To quantify the information quantity, we define the \textit{text length} as the total count of characters including whitespace and punctuation, as these elements occupy physical dimensions during the rendering process, thereby directly determining the information quantity of the image.

When plotting the ED against the text length, a distinct pattern emerges, as illustrated in Figure~\ref{fig:scatter_plot_main}. The performance does not degrade linearly with increasing information quantity; instead, the relationship between ED and text length exhibits three distinct regimes. In the initial regime of short text lengths, the model exhibits a \textbf{Stable Phase} where the ED remains near-zero. Subsequently, the behavior shifts to the \textbf{Zone I: Instability Phase}. In this region, the average ED begins to rise, accompanied by significant instability; for identical text lengths, the ED fluctuates drastically, ranging from low error to substantial error. Finally, once the text length surpasses a threshold, the model crosses a boundary we term the ``Hard Wall'' and enters the \textbf{Zone II: Collapse Phase}. Here, the ED abruptly surges above $0.6$, indicating that the model has lost the capability to reconstruct the image content. Furthermore, across all tested resolutions, we observe a consistent trend: higher resolutions extend the range of the Stable Phase and shift the Hard Wall further to the right.

\subsection{Zone I: Spatial Alignment Sensitivity}
\label{sec:spatial_alignment}


A key question arises: what mechanisms drive the divergence between Zone I and Zone II? We first investigate the origins of the high variance observed in Zone I. We hypothesize that this fluctuation originates from the Spatial Alignment Sensitivity inherent to ViTs~\cite{rojas2024making}. Since ViTs process images by dividing them into fixed-size patches, the model's recognition performance is determined by its ability to integrate features distributed across different patches.

To validate this hypothesis, we designed a \textit{Pixel-Shift Perturbation} experiment. We first rescaled the original images to a size of $R - 16$ (where $R$ denotes the model input resolution and 16 corresponds to the ViT patch size in DeepSeek-OCR) and placed them onto the $R \times R$ canvas. By shifting the starting coordinates $(x,y)$ from $(0,0)$ to $(16,16)$ with a stride of 2 pixels, we traversed the spatial offset of a the ViT patch, generating a set of perturbed variations for each sample. We then recorded the \textit{minimum} ED achievable across all variations. This experiment was conducted on high-ED \textit{Novels} samples spanning Zones I and II, with Group A at $R=640$ and Group B at $R=1024$.

The results, visualized in Figure~\ref{fig:pixel_shift}, reveal a distinct contrast in how these two zones respond to spatial perturbation. For samples located in Zone I, adjusting the spatial alignment results in a restoration of performance. The scatter plot demonstrates that for nearly all high-error samples in this region, the minimum ED drops to near-zero ($<0.05$) merely by shifting the image pixels. Conversely, for samples in Zone II, the error remains high regardless of spatial positioning.

\noindent\textbf{Mechanism Analysis:} Due to the inherent sensitivity of ViTs to spatial alignment, grid partitioning inevitably causes characters to be fragmented across multiple patches. As the information quantity increases, the information load carried by each vision token rises correspondingly. This limits the model's ability to reconstruct these fragmented features. In contrast, within the Stable Phase, the relatively low information quantity grants the model high tolerance to grid misalignment, resulting in minimal reconstruction errors. The Pixel-Shift experiment confirms that Zone I degradation is reversible, as textual content can be recovered provided that a suitable spatial alignment is identified. This indicates that in Zone I, the information is not lost but rather inaccessible due to misalignment.

\subsection{Zone II: Legibility vs. Capacity}
\label{sec:legibility_vs_capacity}

\begin{figure}[t]
\centering
\includegraphics[width=0.9\columnwidth]{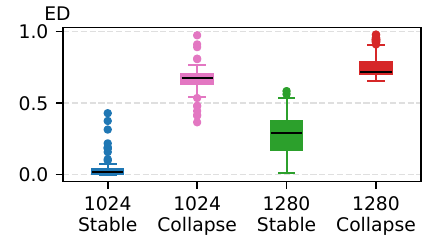}
\caption{Visual density alignment on Novels: 1024 Stable denotes Stable-group samples (TL < 5k) evaluated at input resolution $R=1024$ (analogously for other labels). Despite matched character scale, Collapse-group samples (TL > 15k) retain high ED, supporting an information-capacity limit.}
\label{fig:canvas_experiment}
\end{figure}

To determine the underlying mechanism driving the collapse in Zone II, we propose two potential hypothesis. \textit{Hypothesis A (Visual Legibility Limit)} suggests that the failure is perceptual, positing that as text length increases, the image dimensions expand, causing characters to shrink below the recognizable threshold when resized to the model's input resolution. Conversely, \textit{Hypothesis B (Information Capacity Limit)} suggests that while characters remain visually distinguishable, the fixed number of vision tokens is insufficient to encode the extensive information quantity.

To verify these hypotheses, we designed a controlled experiment using the \textit{Novels} dataset. We established two distinct groups: a \textit{Stable Group}, containing 150 samples with text lengths $< 5,000$ (from the Stable Phase), and a \textit{Collapse Group}, containing 150 samples with text lengths $> 15,000$ (from the Collapse Phase).

We employed a \textit{Visual Density Alignment} strategy to standardize the visual scale. We set the dimensions of a blank canvas to $3584 \times 3584$ (a size sufficient to cover the image size of the longest text in the \textit{Novels} dataset) and pasted the images from both groups onto this fixed canvas. This procedure ensures that when these images are resized to the model's input resolution, the effective character size and pixel density are at an identical level.

The results, illustrated in Figure~\ref{fig:canvas_experiment}, reveal a decisive divergence. The Stable Group maintains better recognition despite the characters being rendered at the exact same visual scale as those in the failure cases. In contrast, the Collapse Group continues to exhibit performance degradation. This observation conclusively refutes the Visual Legibility Hypothesis. Since the model successfully recognizes the Stable Group samples, the collapse in Zone II is not caused by the inability to resolve the text. Instead, it validates the hypothesis that the collapse in Zone II is triggered when the text quantity exceeds the capacity limit.

\section{Scaling Law of Recognition Limits}
\label{sec:modeling_interplay}

Conclusions drawn from Section~\ref{sec:hard_wall_mechanics} suggest that the model's performance is driven by the interaction between information quantity and vision token count. To examine the impact of typographic density, we conducted a comprehensive evaluation on 192 configurations comprising 4 resolutions and 48 layout settings from the \textit{Novels} dataset. This analysis revealed that higher typographic density correlates with increased recognition error. Additional experimental details are provided in the Appendix~\ref{app:layout_impact}. Having identified these as the three critical determinants, we explicitly incorporate these variables into a unified probabilistic framework to quantitatively define the interplay among them.

\subsection{Formulation of Metrics}
We first define two critical variables: the average vision token load and the visual density.

\noindent\textbf{1. Average Vision Token Load ($G$):} This metric measures the average information quantity that each vision token is required to encode. Let $N_{char}$ be the total text length and $N_{token}$ be the number of vision tokens available at the current input resolution $R$. We define: $G = N_{char}/{N_{token}}$. A higher $G$ implies that each vision token bears a heavier semantic payload.

\noindent\textbf{2. Visual Density ($V$):} To capture the impact of layout, we define $V$ as the average number of text lines contained within a single ViT patch. Let $S$ denote the patch size. For a specific typographic configuration, let $w_f$ and $h_f$ be the pixel width and height of a single character, with character spacing $c_s$ and line spacing $l_s$. 

Assuming the generated image has an aspect ratio of $k$ (width/height) and contains $n_{line}$ characters per line on average, the geometric relationship implies:
\begin{equation}
    n_{line} \cdot (w_f + c_s) = k \cdot \frac{N_{char}}{n_{line}} \cdot (h_f + l_s)
\end{equation}
When the image is resized to the model's input resolution $R$, the vertical resizing scale $r_{height}$ is given by $R / [\frac{N_{char}}{n_{line}}(h_f + l_s)]$. Consequently, the number of text lines captured within a single patch height $S$ is calculated as:
\begin{equation}
    V = \frac{S}{r_{height} \cdot (h_f + l_s)} = \frac{S}{R} \sqrt{\frac{N_{char}(w_f + c_s)}{k(h_f + l_s)}}
\end{equation}
This formulation explicitly captures how layout parameters influence the effective pixel density perceived by the model after resizing. A higher $V$ signifies a greater vertical density of text lines within a local patch.

\subsection{Modeling Phase Transitions via Latent Difficulty}
We hypothesize that the sequential transition from the Stable Phase through the Instability Phase to the Collapse Phase represents a probabilistic process dictated by a unified difficulty metric.

We propose a \textit{Latent Difficulty ($Z$)} constructed as a log-linear combination of the visual density ($V$) and token load ($G$):
\begin{equation}
\label{eq:scaling_law}
    Z = w_0 + a \log V + \alpha \log G
\end{equation}
Here, $a$ and $\alpha$ are learnable scaling exponents representing the model's universal sensitivity to density and capacity, respectively, while $w_0$ is a resolution-specific bias term.

Based on the empirical distribution of the ED, we introduce a binary latent variable $z \in \{0, 1\}$, where $z=0$ represents the \textit{Stable Mechanism} and $z=1$ represents the \textit{Collapse Mechanism}. The expected ED is modeled as a mixture model:
\begin{equation}
    p[D|Z] = (1 - \pi(Z)) \cdot p_0(D) + \pi(Z) \cdot p_1(D)
\end{equation}
where $D$ denotes the edit distance. The probability of entering the collapse mechanism is defined by a sigmoid function of the latent difficulty: $\pi(Z) = \sigma(Z)$. We adopt Beta distributions for each mechanism: $D \mid (z=k) \sim \text{Beta}(\alpha_k, \beta_k)$.

\subsection{Model Fitting}

\begin{figure}[t]
\centering
\includegraphics[width=\columnwidth]{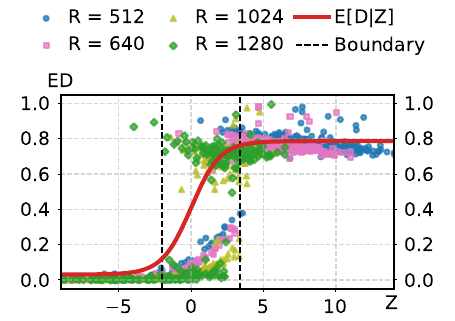}
\caption{\textbf{Alignment of performance curves in the Latent Difficulty space.} The alignment of data points from different resolutions confirms the scaling law. The vertical boundaries show that the transition thresholds into Zone I and Zone II are identical across all configurations when plotted against $Z$.}
\label{fig:universal_scaling}
\end{figure}

We estimated the model parameters using the empirical results detailed in Appendix~\ref{app:layout_impact}. We employed the Expectation-Maximization (EM) algorithm to maximize the log-likelihood. 
In the \textit{E-Step}, we calculate the posterior probability $r_i$ that sample $i$ belongs to the collapse regime. 
In the \textit{M-Step}, we update the scaling parameters $(w_0, a, \alpha)$ via weighted logistic regression using $r_i$ as soft labels, and simultaneously update the Beta parameters via weighted Maximum Likelihood Estimation (MLE).

Crucially, we enforce the scaling laws $a$ and $\alpha$, as well as the Beta parameters $\beta_0, \beta_1$, to be \textbf{shared} across all resolutions. This constraint forces the model to learn universal physical laws, while allowing $w_0$ to capture resolution-dependent baselines. A specific layout configuration (Font 28, Line Spacing 6, Char Spacing 0) was held out entirely from the training set to validate generalization.

\paragraph{Results Analysis.} The fitted parameters are shown in Table~\ref{tab:scaling_params}. Based on the fitted mixture model parameters, we derive the theoretical performance boundaries. The expected edit distance is formulated as:
\begin{equation}
    \mathbb{E}[\text{D}|Z] = (1-\pi(Z))\mu_0 + \pi(Z)\mu_1
\end{equation}
where $\mu_0 = 0.03$ and $\mu_1 = 0.79$ are the means of the success and failure Beta distributions. Given an ED threshold $e^*$, the corresponding $Z^*$ is obtained via $\pi^* = (e^* - \mu_0)/(\mu_1 - \mu_0)$ and $Z^* = \log(\pi^*/(1-\pi^*))$.

We define phase boundaries using expected edit distance thresholds:
\begin{itemize}
    \item \textbf{Stable}: $\mathbb{E}[D] < 0.1$ ($Z < -2.3$)
    \item \textbf{Instability}: $0.1 \leq \mathbb{E}[D] \leq 0.7$
    \item \textbf{Collapse}: $\mathbb{E}[D] > 0.7$ ($Z > 2.0$)
\end{itemize}

\begin{table}[t]
\centering
\caption{Fitted parameters of the scaling law model.}
\label{tab:scaling_params}
\begin{tabular}{cc|cc}
\toprule
\multicolumn{2}{c|}{Shared} & \multicolumn{2}{c}{Resolution-specific} \\
\midrule
$a$ & 2.91 & $w_0^{(512)}$ & -45.57 \\
$\alpha$ & 5.53 & $w_0^{(640)}$ & -46.14 \\
$\alpha_0$ & 0.43 & $w_0^{(1024)}$ & -46.23 \\
$\beta_0$ & 13.87 & $w_0^{(1280)}$ & -42.99 \\
$\alpha_1$ & 13.27 & & \\
$\beta_1$ & 3.59 & & \\
\bottomrule
\end{tabular}
\end{table}

The critical text lengths at phase boundaries are:
\begin{equation}
    N_{\text{boundary}}^{(R)} = N_{\text{token}}^{(R)} \cdot \exp\left(\frac{Z^* - w_0^{(R)} - a\log V}{\alpha}\right)
\end{equation}
where $Z^* = -2.3$ for stable and $Z^* = 2.0$ for collapse.


\paragraph{Key Findings.} Our analysis reveals two critical insights:
\begin{enumerate}
    \item \textbf{Universal transition width}: The ratio $N_{\text{collapse}}/N_{\text{stable}} = \exp(4.3/\alpha) \approx 2.2$ is resolution-independent. The Instability phase always spans a $2\times$ range in text length.

    \item \textbf{Load dominates density}: The coefficient $\alpha = 5.53$ (load $G$) exceeds $a = 2.91$ (density $V$) by $1.9\times$, indicating that total text length matters more than local character density for predicting failure.
\end{enumerate}
We also visualize the performance of the fitted model on the validation set, as shown in Figure~\ref{fig:universal_scaling}. By projecting diverse configurations into the $Z$ space, we demonstrate that the model's failure modes follow a universal scaling law.


\paragraph{Generalization across VLM Architectures}

To verify the generalizability of our probabilistic model, we applied the Latent Difficulty $Z$ derived in Section~\ref{sec:modeling_interplay}. Crucially, we \textbf{froze} the scaling exponents $a$ and $\alpha$ to the values learned from DeepSeek-OCR, and only re-estimated the resolution-specific bias term $w_0$ for InternVL3.5-8B and Qwen2.5-VL-8B. We observed that both models exhibit the identical three-phase transition pattern observed in DeepSeek-OCR. The fact that the scaling laws transfer across different architectures despite differences in training data and architectural designs suggests that the coefficients $a$ and $\alpha$ capture the capacity constraints of visual tokens. Due to space constraints, detailed results and discussion are provided in Appendix~\ref{app:generalization}.


\section{Discussion and Implications}

\paragraph{Spatial Sensitivity in ViT Architectures} The identified \textit{Instability Phase} highlights a structural weakness in ViT: the rigidity of grid-based patch partitioning. Our pixel-shift experiments demonstrated that information is often preserved but becomes inaccessible due to misalignment between semantic boundaries and patch borders. This implies that standard ViT architectures are suboptimal for dense text compression, as they lack the shift-invariance inherent in CNNs or the flexibility of sliding windows. 

\paragraph{Towards Compression-Aware VLM Design} The probabilistic scaling law proposed in Eq.~\ref{eq:scaling_law} provides a practical compass for VLM efficiency. By unifying visual density ($V$) and average vision token load ($G$) into a single \textit{Latent Difficulty} metric ($Z$), we can now predict the minimal token budget required for a given document without running expensive forward passes. This enables an adaptive inference paradigm: simpler documents can be processed with aggressive compression, while visually dense documents can dynamically trigger higher resolutions or tiling strategies. Such "content-aware" computation is essential for deploying long-context VLMs in resource-constrained environments.

\section{Conclusion}
\label{sec:conclusion}

In this paper, we presented the first systematic study on the information capacity of vision tokens. Through controlled experiments on dense text images, we identified a phase-transition phenomenon characterized by three distinct regimes: a Stable Phase, an Instability Phase driven by spatial misalignment, and a Collapse Phase triggered by capacity exhaustion. Furthermore, we distinguished the mechanical origins of these failures, separating reversible patch-alignment errors from irreversible information loss. Based on these insights, we formulated a probabilistic scaling law that accurately predicts recognition limits by modeling the interaction between visual density and average vision token load. Our work establishes a theoretical boundary for vision-to-text compression, suggesting that while vision tokens offer a promising path for context compression, they are bound by intrinsic information limits that must be explicitly modeled and optimized in future VLM architectures.

\section{Limitations}

While our study offers significant insights, it is subject to several limitations that outline directions for future work:

\paragraph{Model Family Coverage}
While we validated the generalizability of our findings across several representative VLM architectures, our conclusions are most directly applicable to modern ViT-based vision tokenizers. Architectures employing fundamentally different tokenization or recognition mechanisms may exhibit distinct transition behaviors not captured by our current framework.

\paragraph{Linguistic Diversity}
Our data synthesis pipeline primarily utilized English text. However, other languages (such as Chinese or Japanese) possess significantly higher information density per character and distinct structural characteristics. We hypothesize that the position of the "Hard Wall" (capacity limit) may shift for these scripts, necessitating a recalibration of the scaling parameters to account for varying varying linguistic densities.

\paragraph{Mitigation Strategies}
While this study successfully diagnosed the mechanisms of alignment sensitivity and capacity exhaustion, it did not systematically explore countermeasures. Specifically, we did not investigate whether specific architectural interventions could mitigate the performance degradation caused by alignment sensitivity, nor did we evaluate potential schemes to fundamentally enhance the intrinsic information capacity of vision tokens. Addressing these optimization challenges remains a critical avenue for future research.
\bibliography{temp}

\appendix

\section{Data Synthesis Details}
\label{app:data_synthesis}

This appendix details the technical implementation of our data synthesis pipeline, including the image rendering algorithm and the specific segmentation strategies employed for each semantic domain.

\subsection{Dataset Selection and Segmentation}
To ensure the "Block-wise Shuffling" strategy is effective, we standardized the block size across all domains. The \textit{Novels} dataset serves as the baseline for token length distribution.

\paragraph{1. Novels}
We selected Jane Austen's \textit{Pride and Prejudice} (via Project Gutenberg) as the primary source for the narrative domain. The raw text was first segmented by the \texttt{CHAPTER} keyword. To establish a baseline for expected token length across all datasets, each chapter was further partitioned into four discrete text blocks.

\paragraph{2. Legal Statutes}
For the legal domain, we utilized \textit{Postconviction Remedies} from the CALI Library\footnote{\url{https://www.cali.org/books/postconviction-remedies}}, converting the original DOCX documents into plain text. Unlike the novel dataset, legal chapters exhibited drastic variations in length. To normalize this and ensure distribution consistency with the \textit{Pride and Prejudice} baseline, we implemented a hierarchical segmentation strategy. We defined the length of the first chapter as a standard unit ($L_{std}$) and subdivided longer chapters into multiples of this unit. Each resulting sub-chapter was then split into six "pages," which were finally divided into four blocks each.

\paragraph{3. Economics}
The Economics dataset is derived from Adam Smith's seminal work, \textit{An Inquiry into the Nature and Causes of the Wealth of Nations}\footnote{\url{https://www.gutenberg.org/ebooks/3300}}. Following the established preprocessing pipeline, the text was segmented into blocks with token counts strictly aligned with the baseline size.

\paragraph{4. Medicine}
Medical texts were sourced from \textit{Anomalies and Curiosities of Medicine} by George M. Gould and Walter L. Pyle\footnote{\url{https://www.gutenberg.org/ebooks/747}}. Similar to the previous domains, the content was processed and segmented into discrete blocks matching the standardized baseline dimensions.

\paragraph{5. Newspapers}
To represent journalistic text, we compiled individual news stories from \textit{Daily Stories of Pennsylvania} by Frederic A. Godcharles\footnote{\url{https://www.gutenberg.org/ebooks/69956}}. These stories were concatenated into a single stream and subsequently re-segmented into blocks that adhere to the control group's size constraints.

\paragraph{6. Personal Letters}
The epistolary dataset was constructed using \textit{Letters from a Self-Made Merchant to His Son} by George Horace Lorimer\footnote{\url{https://www.gutenberg.org/ebooks/21959}}. The letters were individually processed and segmented into blocks consistent with the baseline token distribution.

\subsection{Image Rendering Pipeline}
The synthesized text is rendered into images using the \textbf{DejaVuSans} font. To ensure the synthesized images meet the requirements of square aspect ratios and uniform text density, we implemented a multi-stage rendering engine using Python and the Python Imaging Library (PIL). The specific procedure is as follows:

\paragraph{1. Text Preprocessing}
Raw text data is first sanitized to remove encoding artifacts. We strip leading and trailing whitespace and merge consecutive space characters into a single space. To retain paragraph structure within the visual block, sequences of multiple spaces in the source text are treated as paragraph delimiters, introducing a line break in the rendering process.

\paragraph{2. Iterative Layout Optimization}
Since the text length varies significantly across samples, fixing the image width would result in extreme aspect ratios (e.g., long vertical strips). We employ an iterative algorithm to determine the optimal canvas dimensions:
\begin{enumerate}
    \item \textbf{Initial Estimation:} Based on the total text length, average character width (of the \textit{DejaVuSans} font), and total line height (font size + line spacing), we estimate an initial wrapping width $W_{init}$ intended to produce a square image.
    \item \textbf{Trial Rendering:} The text is logically wrapped using $W_{init}$ to calculate the resulting image height $H$.
    \item \textbf{Aspect Ratio Check:} We calculate the aspect ratio $AR = H / W$. The target range is defined as $0.9 \le AR \le 1.1$.
    \item \textbf{Adjustment Loop:} If $AR$ falls outside the target range, $W$ is adjusted iteratively. If $AR > 1.1$ (too tall), $W$ is increased; if $AR < 0.9$ (too wide), $W$ is decreased. This process repeats until the constraint is met or a maximum iteration limit is reached, selecting the layout closest to a square.
\end{enumerate}

\paragraph{3. Typographical Rendering}
Once the layout is finalized, the text is rendered onto a white background with the following specifications:
\begin{itemize}
    \item \textbf{Vertical Spacing:} The height of each line is determined by the sum of the font size and the specified line spacing parameter.
    \item \textbf{Horizontal Spacing:} A fixed character spacing parameter is added between individual characters to modify density.
\end{itemize}

\paragraph{4. Output Generation}
The final canvas is exported as a JPEG image. We explicitly set the DPI and compression quality to minimize artifacts that could interfere with the OCR process.

\section{Impact of Visual Layout}
\label{app:layout_impact}
To investigate how visual layout parameters affect the model's recognition performance, we conducted experiments varying font size, line spacing, and character spacing. We select \textit{Novels} as the test domain and utilize the deepseek-ocr model at resolutions 512, 640, 1024, and 1280. The font sizes tested were 20, 28, and 36 pixels. Line spacing values included 0, 6, 24, and 42 pixels, while character spacing values were set to -1, 0, and 7 pixels. There are a total of 192 unique combinations of layout parameters. 

\begin{figure*}[t]
\centering
\includegraphics[width=\textwidth]{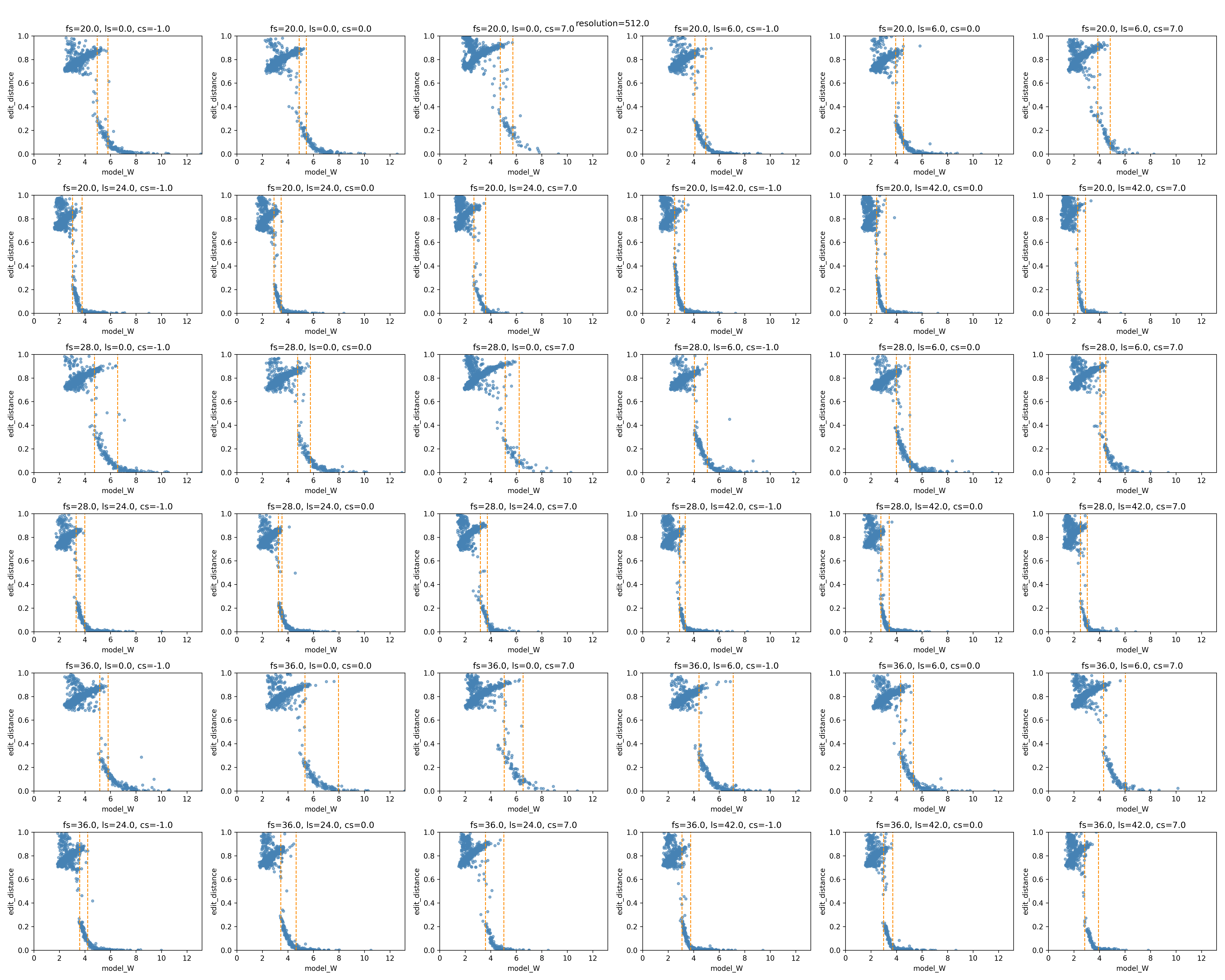}
\caption{Scatter plots of edit distance versus character pixel size for each layout configurations in 512 resolution.}
\label{fig:512_MW}
\end{figure*}

\begin{figure*}[t]
\centering
\includegraphics[width=\textwidth]{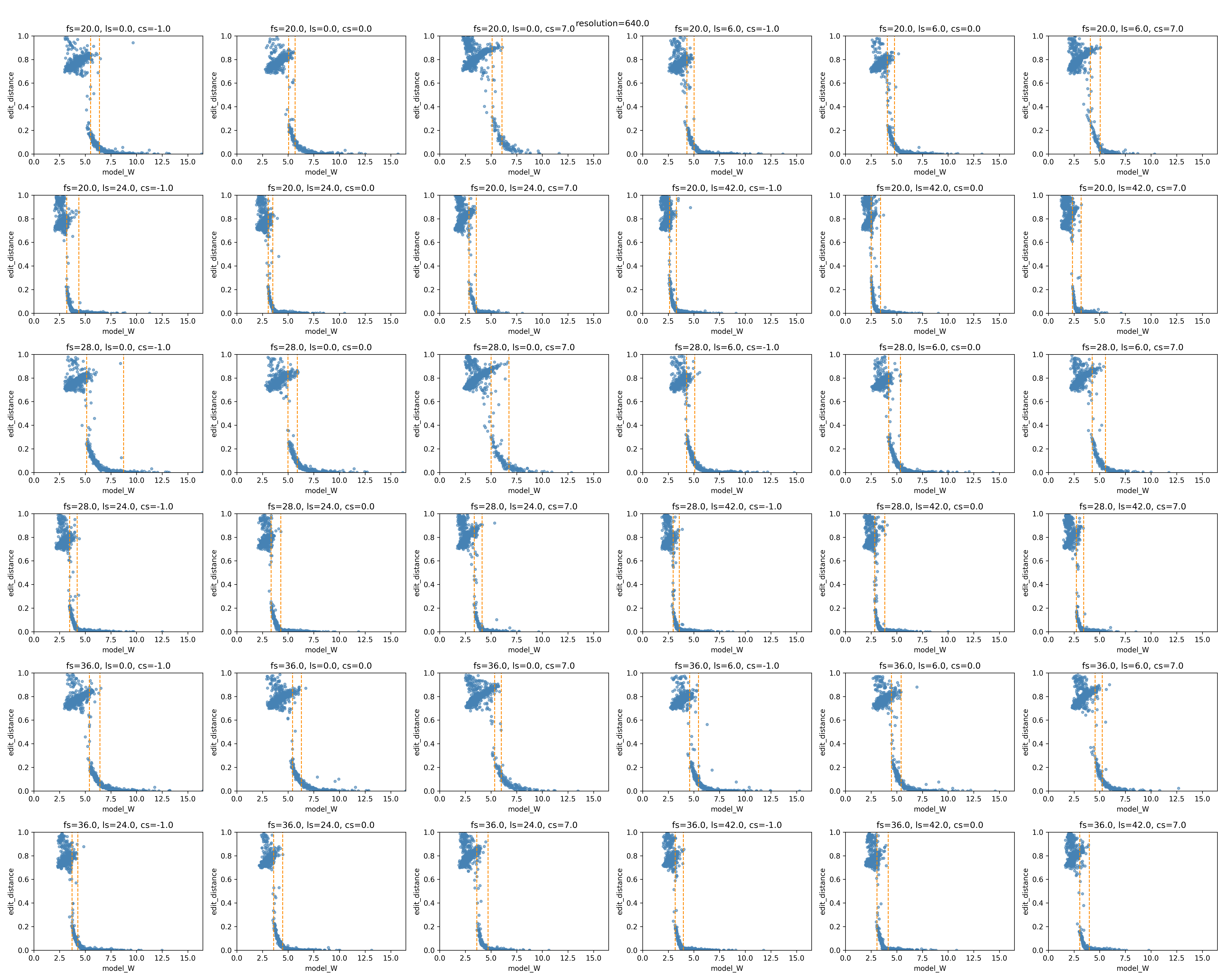}
\caption{Scatter plots of edit distance versus character pixel size for each layout configurations in 640 resolution.}
\label{fig:640_MW}
\end{figure*}

\begin{figure*}[t]
\centering
\includegraphics[width=\textwidth]{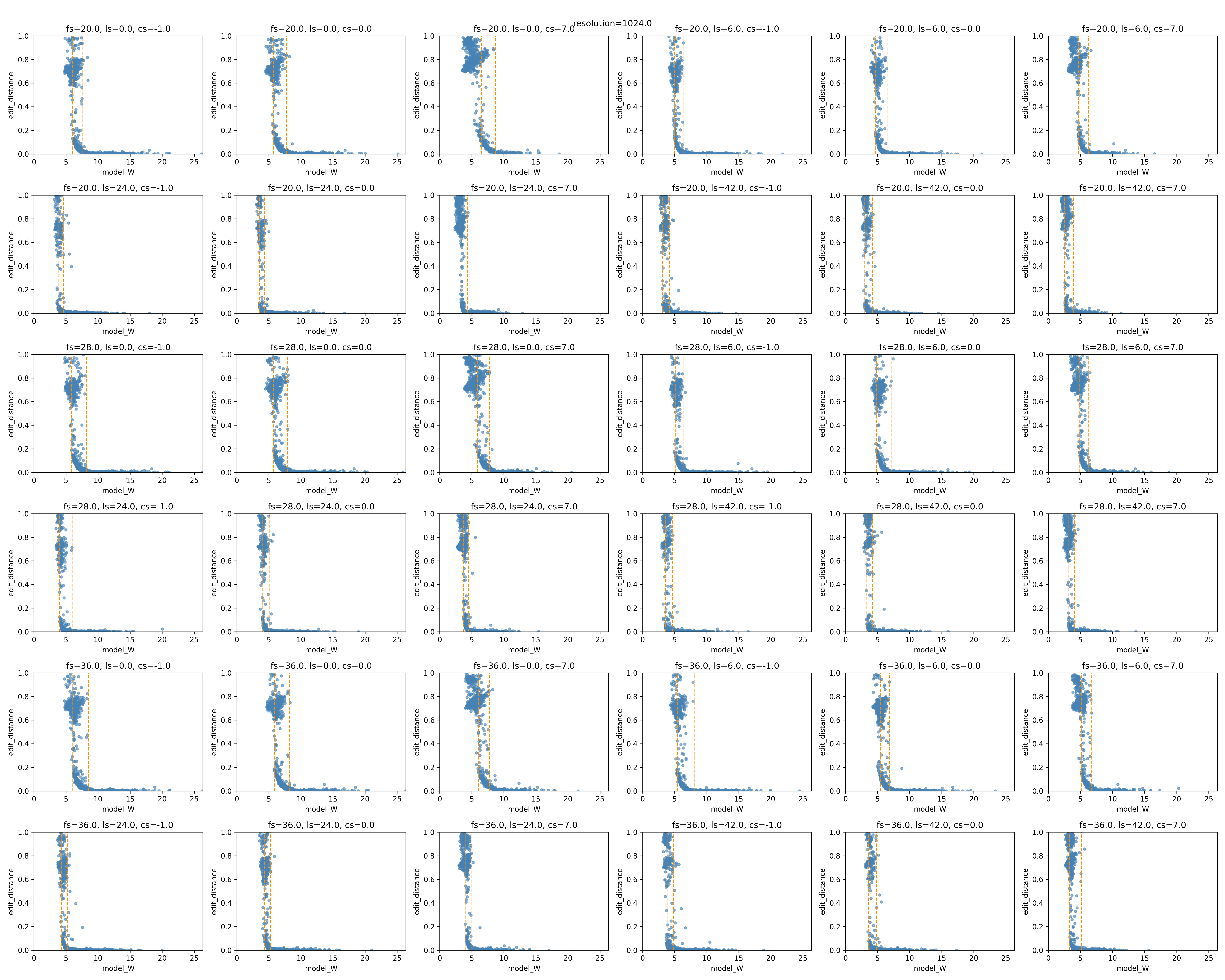}
\caption{Scatter plots of edit distance versus character pixel size for each layout configurations in 1024 resolution.}
\label{fig:1024_MW}
\end{figure*}

\begin{figure*}[t]
\centering
\includegraphics[width=\textwidth]{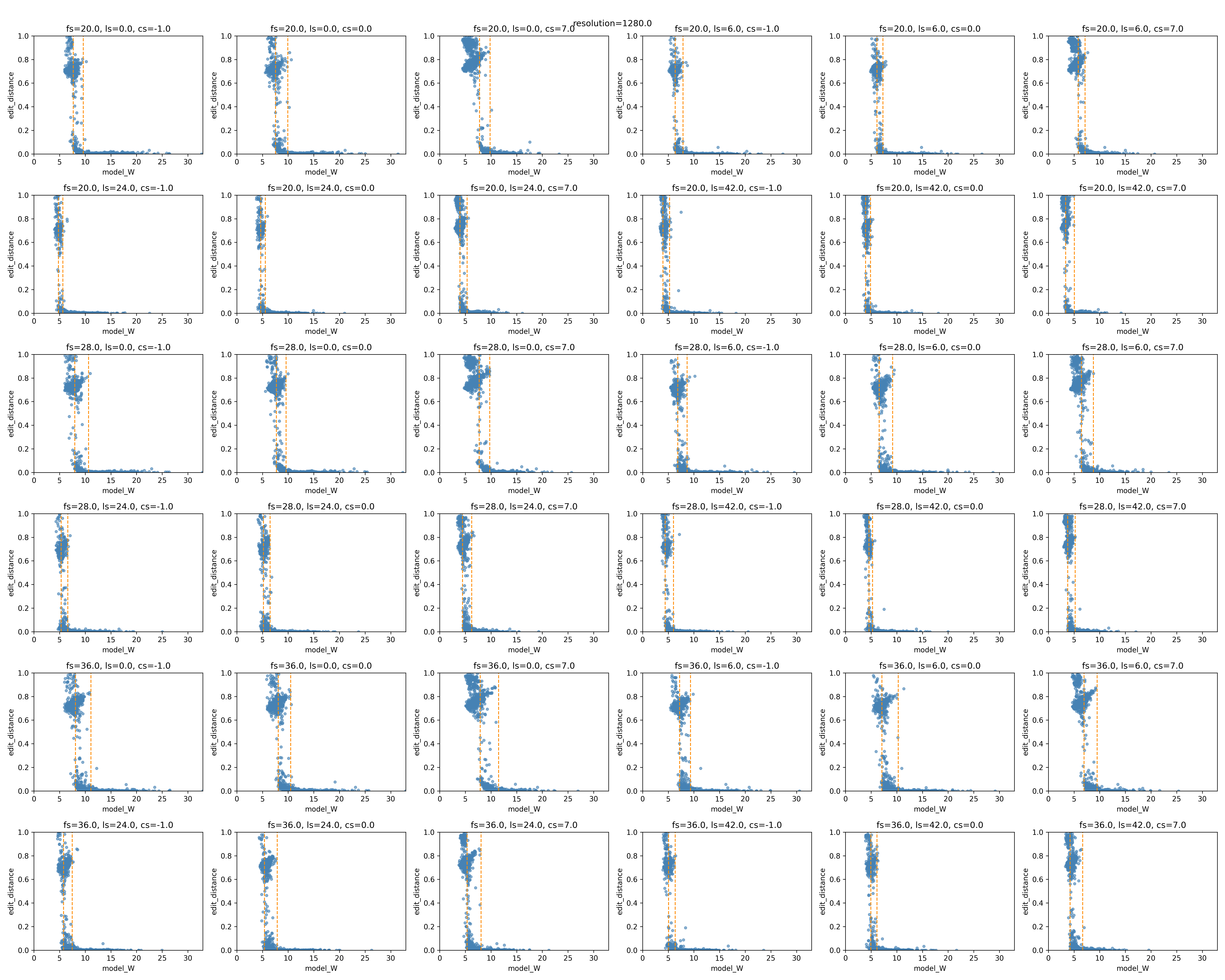}
\caption{Scatter plots of edit distance versus character pixel size for each layout configurations in 1280 resolution.}
\label{fig:1280_MW}
\end{figure*}

\paragraph{Observations.} Scatter plots of edit distance versus character pixel size (after resizing) for each resolution are presented in Figure \ref{fig:512_MW}, \ref{fig:640_MW}, \ref{fig:1024_MW}, \ref{fig:1280_MW}. At the same resolution, with the horizontal axis representing the pixel size of characters as seen by the model (after resizing) and the vertical axis representing edit distance, we observe that larger line spacing leads to smaller character pixel sizes at the critical points where the model enters Zone I and Zone II. This confirms that density indeed affects performance.

At the same resolution, when the model observes characters of the same pixel size, larger font sizes result in larger generated images under the same line spacing and character spacing settings, leading to a larger resize scale. This causes larger font sizes to have smaller line spacing (as perceived by the model) when the model sees characters of the same pixel size, thus increasing density. Therefore, larger font sizes correspond to larger character pixel sizes at the critical points of entering Zone I and Zone II. So we can conclude that font size affects performance through its impact on density.

\begin{figure*}[t]
\centering
\includegraphics[width=\textwidth]{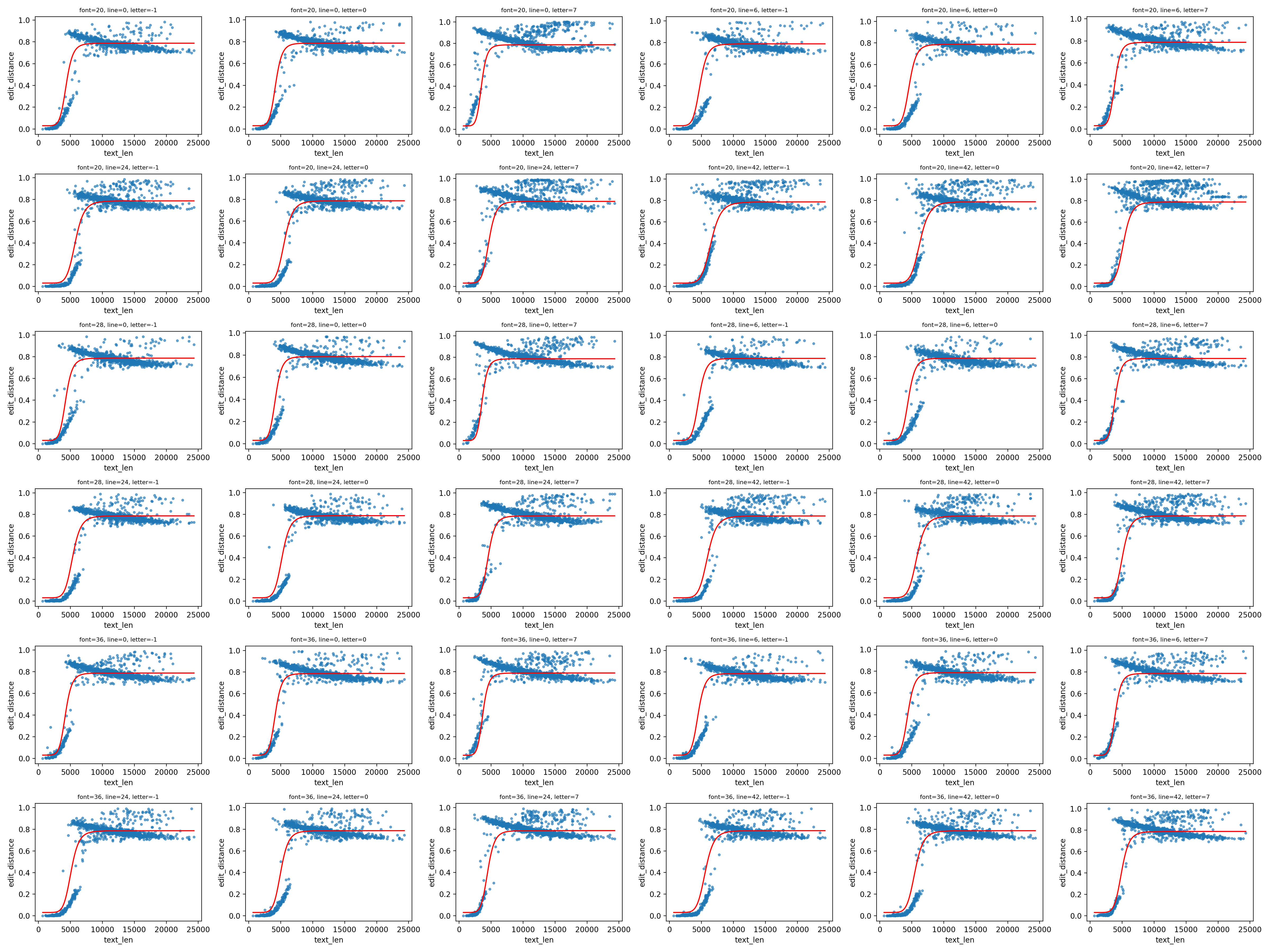}
\caption{Scatter plots of edit distance versus text length for each layout configurations in 512 resolution.}
\label{fig:512_TL}
\end{figure*}

\begin{figure*}[t]
\centering
\includegraphics[width=\textwidth]{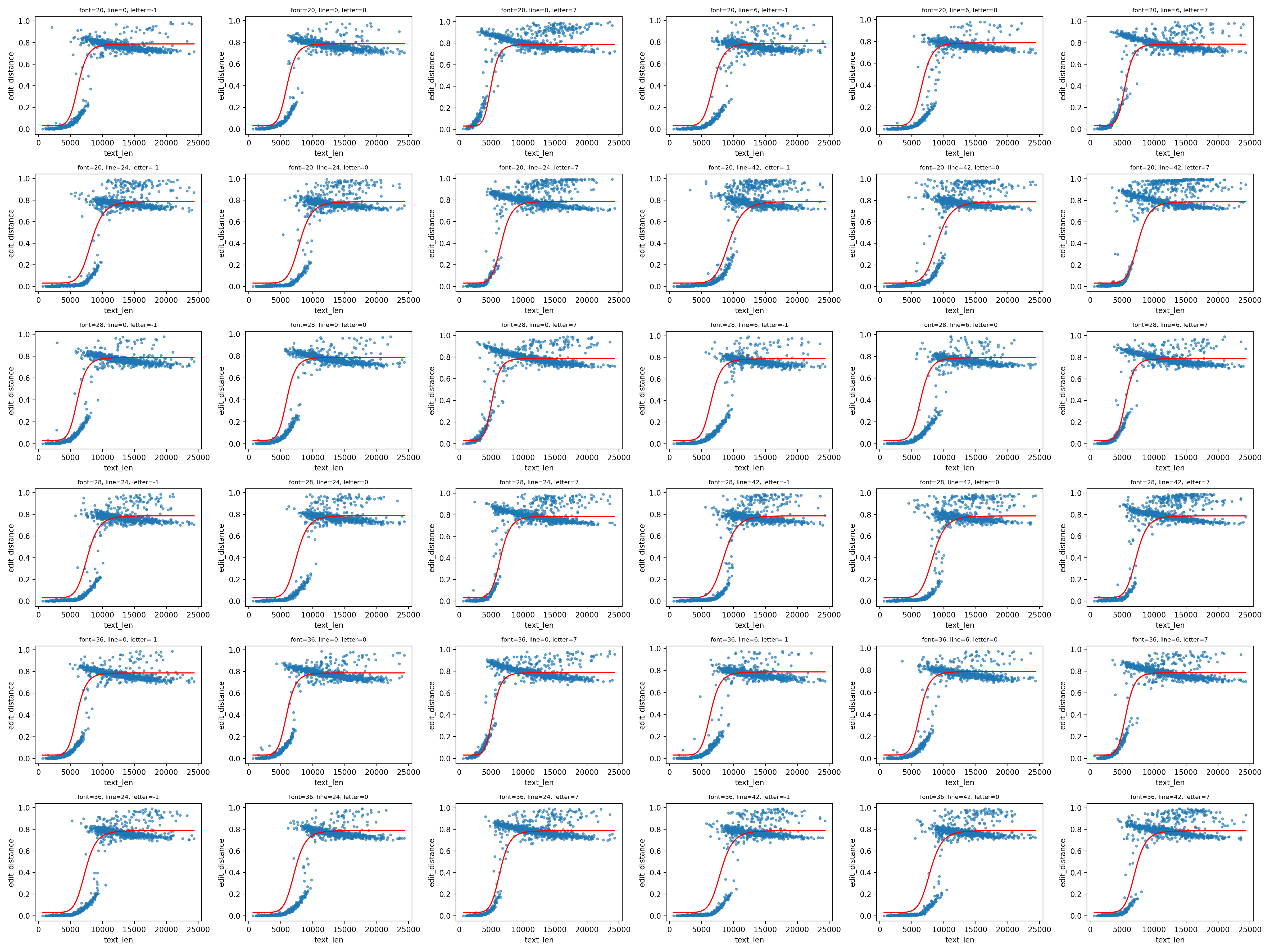}
\caption{Scatter plots of edit distance versus text length for each layout configurations in 640 resolution.}
\label{fig:640_TL}
\end{figure*}

\begin{figure*}[t]
\centering
\includegraphics[width=\textwidth]{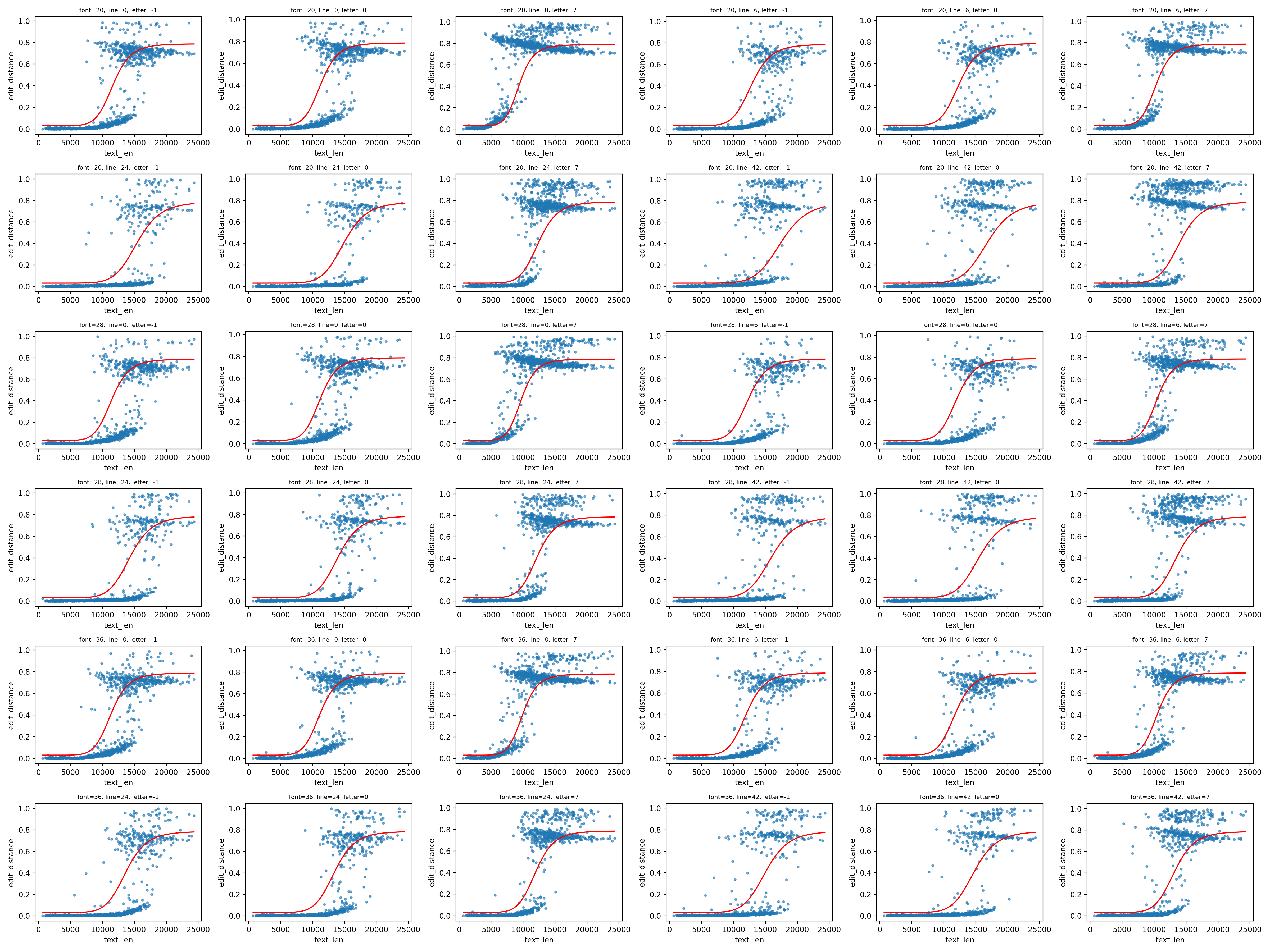}
\caption{Scatter plots of edit distance versus text length for each layout configurations in 1024 resolution.}
\label{fig:1024_TL}
\end{figure*}

\begin{figure*}[t]
\centering
\includegraphics[width=\textwidth]{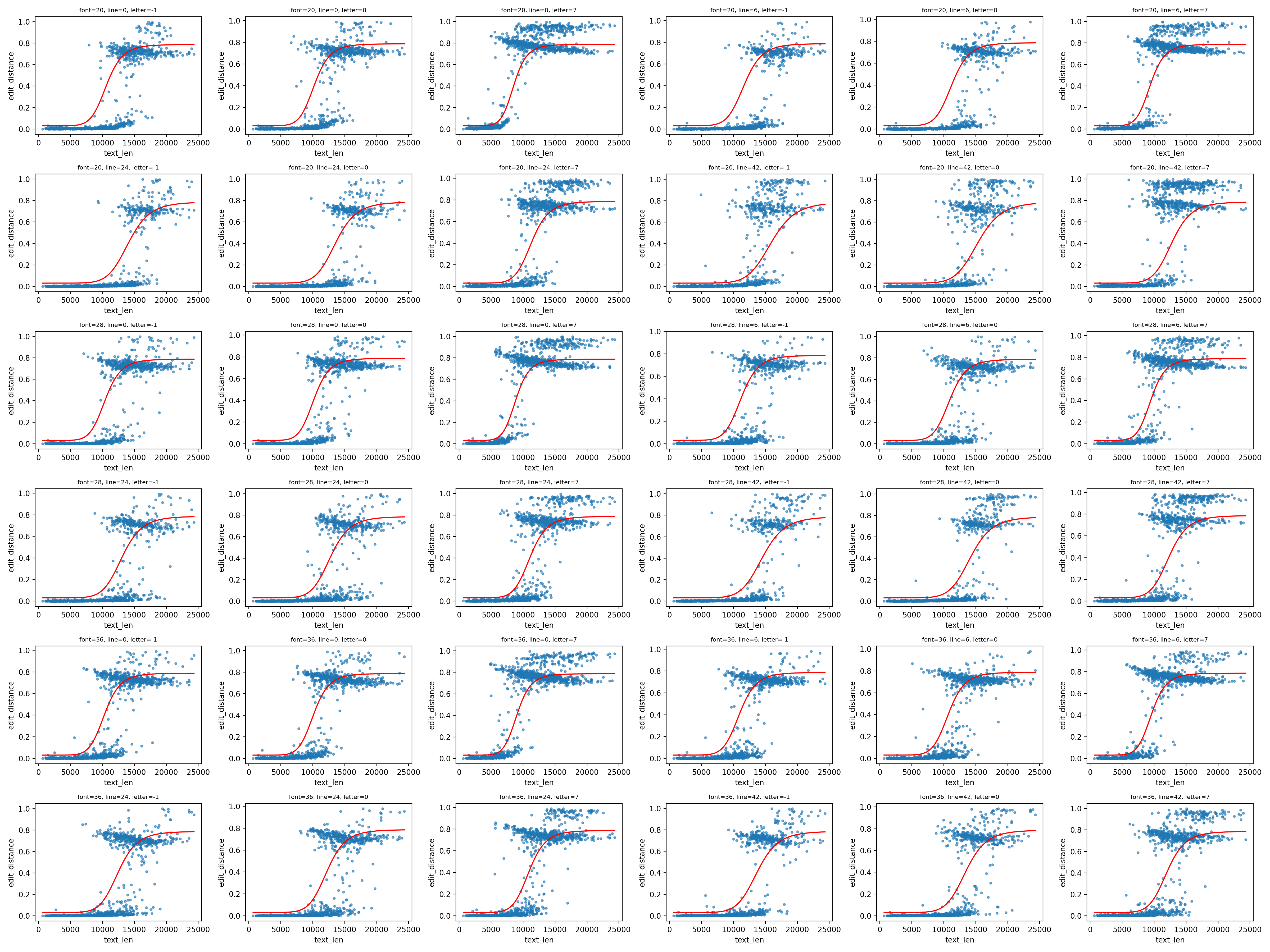}
\caption{Scatter plots of edit distance versus text length for each layout configurations in 1280 resolution.}
\label{fig:1280_TL}
\end{figure*}

\paragraph{Results Fitting.} Scatter plots of edit distance versus text length for each resolution are presented in Figure \ref{fig:512_TL}, \ref{fig:640_TL}, \ref{fig:1024_TL}, \ref{fig:1280_TL}. The red curves represent the fitted relationships based on our proposed difficulty metric $Z = w_0 + a\log V + \alpha \log G$. Across all different layout configurations, the positions where the fitted curves exhibit inflection points are highly consistent with the positions of entering Zone I and Zone II, indicating that the fitted relationship can effectively capture the impact of visual density and average information per vision token on the model's recognition capability.

\section{Generalizability Across Architectures}
\label{app:generalization}

To ascertain whether the phase transition patterns and the identified scaling laws are artifacts of the DeepSeek-OCR architecture or intrinsic properties of Vision Transformers, we extend our evaluation to InternVL3.5-8B and Qwen2.5-VL-8B. Since these models employ dynamic resolution where the vision token count varies adaptively with image dimensions, we conduct two kinds of evaluations. First, we replicate the native dynamic resolution evaluation, allowing each image to be processed at its original size. Second, to facilitate a direct comparison of per-token information capacity requires variable isolation. To isolate the variable of \textit{Average Vision Token Load} ($G$), we intervened in the pre-processing pipeline to enforce a fixed input resolution, thereby locking the vision token budget. Specifically, we resized all images in the \textit{Novel} dataset to $896 \times 896$ for InternVL3.5-8B, resulting in a fixed budget of approximately 1,280 vision tokens. For Qwen2.5-VL-8B, images were resized to $560 \times 560$, yielding approximately 324 vision tokens.

\begin{figure*}[t]
\centering
\includegraphics[width=\textwidth]{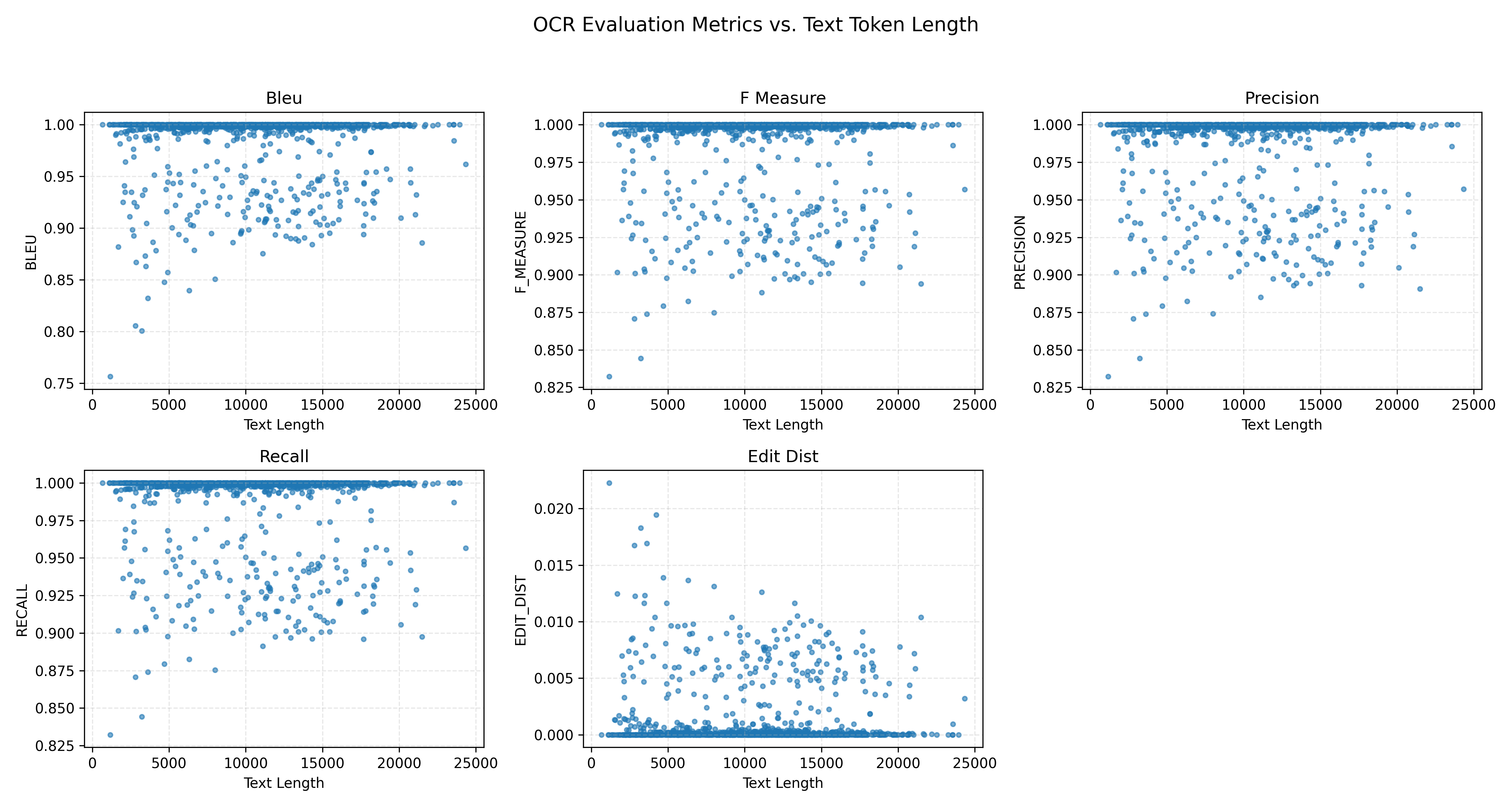}
\caption{Evaluation of Qwen2.5-VL-8B under dynamic resolution.}
\label{fig:qwen_dynamic}
\end{figure*}

\paragraph{Observations.} Regardless of whether dynamic resolution or fixed resolution is employed, the phenomena exhibited by both InternVL3.5-8B and Qwen2.5-VL-8B can be well explained by our proposed probabilistic scaling law. For results under dynamic resolution, as shown in Figure \ref{fig:qwen_dynamic}, the vision tokens used in Qwen2.5-VL-8B increase with the image size. According to our scaling law, the average vision token load $G$ remains relatively stable across different image sizes, that's why the edit distance does not exhibit a significant upward trend as image size increases. In contrast, InternVL3.5-8B also shows the three-phase transition pattern similar to DeepSeek-OCR. 

\begin{figure*}[t]
\centering
\includegraphics[width=0.8\textwidth]{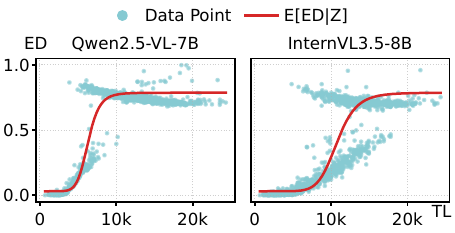}
\caption{Fitting results of InternVL3.5-8B and Qwen2.5-VL-8B under fixed resolution.}
\label{fig:generalization_plots}
\end{figure*}

\paragraph{Fitting Results.} We used the shared parameters $a$, $\alpha$ and the Beta distribution parameters learned from DeepSeek-OCR to fit the data from InternVL3.5-8B and Qwen2.5-VL-8B with fixed image input sizes. The fitting results are shown in Figure \ref{fig:generalization_plots}. From the fitting curves, we can see that the positions of the inflection points in the fitted curves align well with the transitions into Zone I and Zone II across different VLMs, indicating that our proposed scaling law effectively captures the influence of visual density and average information per vision token on the recognition capabilities of various VLM architectures.

\section{Use of AI Tools}
AI-assisted tools were used solely for language polishing and minor grammatical refinement. These tools did not contribute to the formulation of research questions, experimental design, data collection, data analysis, or the generation of scientific claims and conclusions. All technical content, methodological decisions, and interpretations presented in this paper were developed entirely by the human authors.

\end{document}